\documentclass[10pt]{article}
\usepackage[preprint]{tmlr}

\usepackage{amsmath,amssymb,amsthm}
\usepackage{booktabs}
\usepackage{enumitem}
\usepackage{xcolor}
\usepackage[most]{tcolorbox}
\usepackage{paper_style}  
\usepackage{hyperref}
\usepackage{url}

\hypersetup{
    colorlinks=true,
    linkcolor=blue!50!black,
    citecolor=blue!50!black,
    urlcolor=blue!50!black
}

\newtcolorbox{claimbox}[1]{
  colback=pspaper,
  colframe=pscoral,
  title={#1},
  fonttitle=\bfseries,
  boxrule=1pt,
  arc=2pt,
  left=6pt,
  right=6pt,
  top=6pt,
  bottom=6pt
}

\newtheorem{definition}{Definition}

\newcommand{\Target}{\mathcal{O}}
\newcommand{\Features}{\mathcal{E}}
\newcommand{\Process}{\mathcal{P}}
\newcommand{\Conditioning}{\mathcal{C}}

\newcommand{\Score}{\mathcal{S}}

\newcommand{\Contract}{\mathcal{A}}

\newcommand{\given}{\mid}

\title{The Attribution Contract for Generative Language Models}

\author{\name Giang Nguyen \email nguyengiangbkhn@gmail.com \\
        \addr Guide Labs}

\begin{document}

\maketitle

\begin{abstract}
Feature attribution scores each part of an input by how much it explains a model's output. We argue that in generative language models these scores carry no fixed meaning. A classifier has a single output to explain, but a generative model produces its output token by token, and each generated token is both an output and an input, so explaining ``the output'' becomes several distinct questions. We support this claim by proposing the \emph{Attribution Contract}, a framework that names the question a set of attribution scores answers. A contract specifies the model score being explained, what is held fixed, the target output, the generation process, and which features can receive attribution, choices that matter in practice. For example, eligible features decide what a score explains: under a local next-token contract, the model's own generated tokens receive $29\%$ of the attribution mass, comparable to the $25\%$ on the input question, and reading this as a prompt-level explanation is a mistake we name the \emph{self-attribution fallacy}. 
The generative process also shapes feature attribution: on a mixture-of-experts model, Integrated Gradients completeness is unreliable, since recomputing expert routing along the attribution path leaves a large residual that holding the routing fixed substantially lowers,
while a masked-diffusion model, which fixes its choices at generation, has a residual that instead decreases with more steps.
These results show that Attribution Contracts should be carefully specified when proposing and evaluating attribution methods for generative language models, since the meaning and evaluation of an attribution depend on its explanatory setting.
\end{abstract}

\section{Introduction}

Feature attribution methods such as Integrated Gradients~\citep{sundararajan2017axiomatic} often aim to answer a simple question: \emph{which input features contributed to this output?} In the classical setting, this question has a natural shape. A model receives an input $x$, computes a prediction $f(x)$, and an attribution method assigns importance scores to components $x_i$ relative to a scalar target such as a class logit or loss.

In generative language models (LMs), many real-world problems depend on feature attribution. A practitioner debugging a hallucinated summary needs to know whether the hallucinated content came from the source document or from the model's own prior tokens~\citep{ji2023survey, huang2023survey}. A safety researcher auditing a harmful generation needs to know which part of the input triggered it~\citep{ganguli2022redteaming, perez2022redteaming}. A scientist studying in-context learning needs to know which prompt features the model is actually using to make predictions~\citep{min2022rethinking}.

\begin{figure}[!t]
  \centering
  \includegraphics[width=0.80\linewidth]{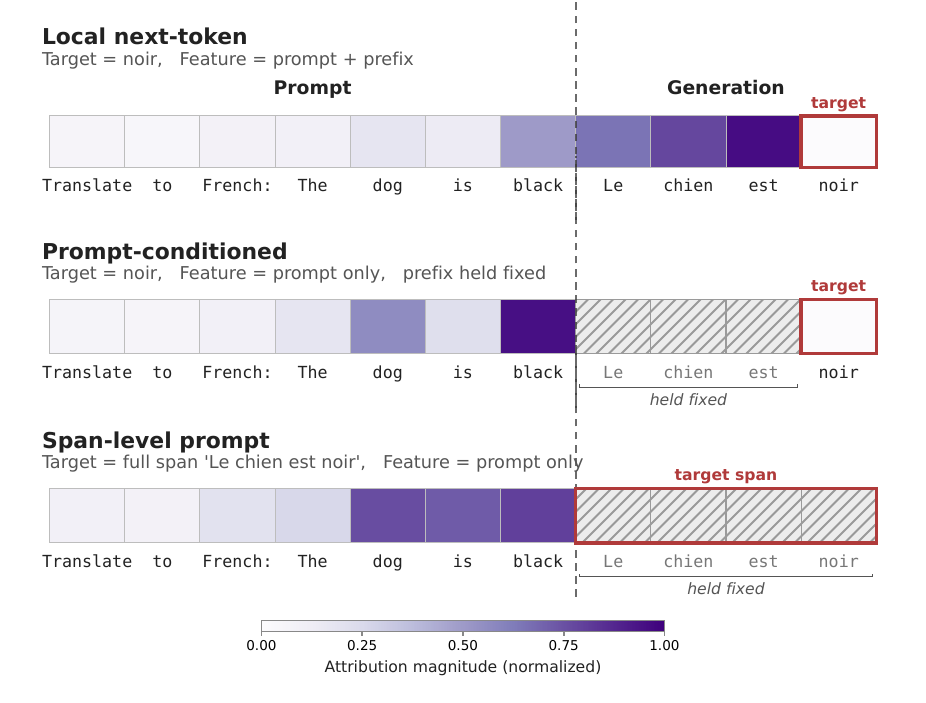}
    \caption{\textbf{The same attribution method produces different attribution maps under different contracts.} All three rows use the same model, prompt, generation, and attribution method (Integrated Gradients~\citep{sundararajan2017axiomatic}). Only the setting differs. \textit{Top:} a local next-token setting attributes the prediction of \texttt{noir} over both prompt and prefix tokens, and mass concentrates on the generated prefix \texttt{Le chien est} because it is predictive of the next token. \textit{Middle:} a prompt-conditioned setting holds the generated prefix fixed (gray, hatched) and attributes only over prompt tokens. ``Held fixed'' means the prefix remains in the forward pass at its actual value but is removed from the attribution path. The result concentrates on \texttt{black}, the prompt token most predictive of \texttt{noir}. \textit{Bottom:} a span-level setting targets the full generated span \texttt{Le chien est noir} and attributes only over prompt tokens; the prefix is again held fixed.}
  \label{fig:attribution-settings}
\end{figure}

Generative LMs break the clean setup of classical attribution. Consider an autoregressive LM with prompt $x$ and generated output $y_{1:T}$:
\begin{equation*}
    p(y_{1:T} \given x)
    = \prod_{t=1}^{T} p(y_t \given x, y_{<t}).
\end{equation*}

If we ask which tokens explain the next-token prediction $y_t$, the previously generated tokens $y_{<t}$ are genuine inputs to the model and may rightly receive high attribution. But those same tokens are also outputs of the model. As a result, feature attribution depends on what exactly we are trying to explain. 
An attribution to earlier-generated tokens may be appropriate if the goal is to explain the model's local prediction of the next token. The same attribution answers a different question if we instead ask how the prompt influenced the overall response $y_{1:T}$. We develop both questions in Section~\ref{sec:settings} and show that they correspond to different feature-attribution contracts.

Unlike in classification settings, the interpretation of a feature-attribution score in generative LMs depends on the explanatory goal.
Suppose a model is prompted:
\begin{quote}
Translate to French: ``The dog is black.''
\end{quote}
and generates:
\begin{quote}
Le chien est noir.
\end{quote}
For the target token ``noir,'' the generated prefix ``Le chien est'' is predictive: it helps determine which token is likely to come next. But if the explanatory question is which input token accounts for the meaning of ``noir,'' the answer should point to ``black,'' \emph{not to the generated prefix}. These are different feature-attribution questions, and the same attribution method produces a different map under each (Figure~\ref{fig:attribution-settings}; quantitative evidence in Section~\ref{sec:feat-flip}).

Attribution to the generated prefix can be correct. Its interpretation depends on the explanatory setting, yet feature-attribution methods for generative LMs often leave this setting implicit~\citep{sarti2023inseq,miglani2023captum,zhao2024reagent,nguyen2024interpretable}. This reflects a broader concern that interpretability claims are underspecified when the question being explained is unclear~\citep{lipton2017mythos}. We call one instance of this problem the \emph{self-attribution fallacy}: treating attribution to generated-prefix tokens as if it answered a prompt-level explanatory question (Section~\ref{subsec:Autoregressive_settings}).

We argue that a feature-attribution score in a generative language model does not carry its own interpretation; rather, the interpretation depends on the explanatory setting. We call such a setting an \emph{Attribution Contract}, and the next section makes it precise and practical.

\paragraph{We summarize the contributions as follows.}
\begin{itemize}[leftmargin=1.2em,itemsep=1pt,topsep=2pt]
\item \textbf{The Attribution Contract.} We specify any feature-attribution claim in a generative model by the tuple $(\Score, \Conditioning, \Target, \Process, \Features)$, spelling SCOPE: the attributed score, what is held fixed, the target output, the generative process, and the eligible features (Section~\ref{sec:attribution_contract}).

\item \textbf{The self-attribution fallacy.} We name the error of reading attribution to a model's own generated tokens as prompt-level explanation (Section~\ref{subsec:Autoregressive_settings}). We show how this can lead to a misleading interpretation: under the local next-token contract, those tokens carry $29\%$ of the attribution mass, comparable to the $25\%$ on the question (Section~\ref{sec:feat-flip}).

\item \textbf{The contract shapes both attribution and evaluation.} Across controlled experiments, changing contract elements substantially changes feature rankings (Sections~\ref{sec:feat-flip} and~\ref{sec:score-flip}) and the behavior of attribution diagnostics (Section~\ref{sec:cond-flip}). We further show that completeness and perturbation-based faithfulness can disagree, so attribution methods should be evaluated relative to the contract they are intended to answer (Section~\ref{sec:eval-flip}).
\end{itemize}

\section{The Attribution Contract}
\label{sec:attribution_contract}

Classical attribution operates on a static schema, $x \to f(x)$, and even there it is sensitive to the explanatory task and evaluation criterion~\citep{bilodeau2024impossibility,nguyen2021effectiveness,adebayo2022post}. Generative language models replace the static schema with a process schema. An autoregressive model conditions each token on the prompt and every earlier token,
\begin{equation*}
    x, y_1, \ldots, y_{t-1} \longrightarrow y_t,
\end{equation*}
and a diffusion model refines a whole sequence through an iterative chain,
\begin{equation*}
    z_T \longrightarrow z_{T-1} \longrightarrow \cdots \longrightarrow z_0 \longrightarrow y,
\end{equation*}
with each $z_t$ a partly resolved sequence~\citep{nie2025llada}.

A generative model makes many predictions, one per step, so the target of an explanation and its eligible features become choices the analyst must make. The Attribution Contract makes those choices explicit.

\begin{definition}[Attribution Contract]

An Attribution Contract, i.e., the explanatory setting under which an attribution claim is made, is a tuple
\begin{equation*}
    \Contract =
    (\Score, \Conditioning, \Target, \Process, \Features)\footnote{Read in order, the tuple spells \emph{SCOPE}, capturing the idea that an Attribution Contract specifies the explanatory scope of a feature-attribution claim.},
\end{equation*}
where
\begin{align*}
    \Score &= \text{model score}, \\
    \Conditioning &= \text{what is held fixed}, \\
    \Target &= \text{output being explained}, \\
    \Process &= \text{generative process}, \\
    \Features &= \text{features eligible to receive attribution}.
\end{align*}
\end{definition}

Each element of the contract plays a distinct role. The score \(\Score\) specifies the model quantity being attributed, such as a logit, probability, loss, or log-likelihood. Two explanations may target the same output while attributing different model scores.

The conditioning \(\Conditioning\) specifies what is held fixed during attribution. Computationally, a fixed variable remains in the forward pass at its actual value but is excluded from the path or perturbation set over which attribution is computed. For example, under Integrated Gradients, a fixed variable is not interpolated along the attribution path. Two contracts that differ only in conditioning can therefore produce different attribution maps over the eligible features.

The target \(\Target\) specifies what is being explained: for example, a next-token prediction, an intermediate diffusion state, or a generated sequence. The process term $\Process$ specifies the structure of generation. 
This is important because feature importance in generative language models often propagates through a sequence of intermediate states rather than in a single input-output step. $\Process$ is always part of the contract, but its computational role varies: under some contracts, attribution is computed by aggregating across the chain; under others, the chain serves only as the conditioning structure of a local prediction. The feature set \(\Features\) specifies which features are eligible to receive attribution. These may include prompt tokens, generated-prefix tokens, intermediate generation states, or other process-related features, depending on the explanatory task.

In the next section, we apply the contract framework to several concrete settings for autoregressive and diffusion language models. These settings show how the same feature-attribution method answers different questions under different contracts.

\section{Feature-attribution settings in generative language models}
\label{sec:settings}

The first row of Table~\ref{tab:contracts} shows the classical classifier setting. Many existing feature-attribution methods for generative LMs are built around local predictive targets, such as next-token probability $p(y_t \given x, y_{<t})$ or sequence log-probability $\log p(y_{1:T} \given x)$ decomposed token by token \citep{sarti2023inseq,miglani2023captum,zhao2024reagent,pramanik2026heta}. The Attribution Contract makes clear that these are only some among many possible feature-attribution settings. In this section, we apply the contract framework to the settings most relevant to generative language models. Table~\ref{tab:contracts} gives all contract elements for each one, and the text below describes the question each setting answers.

\begin{table}[!hb]
\centering
\footnotesize
\setlength{\tabcolsep}{5pt}
\renewcommand{\arraystretch}{1.25}
\caption{Feature-attribution settings organized by Attribution Contract. The classifier row shows the classical setting where attribution is well-defined. Generative LMs have multiple contracts: three autoregressive and three diffusion. Columns follow the SCOPE ordering: $\Score$, $\Conditioning$, $\Target$, $\Process$, $\Features$. A dash in the conditioning column indicates that no variable is held fixed.}
\label{tab:contracts}
\begin{tabular}{@{}llllll@{}}
\toprule
\headrow \textbf{Setting} & $\Score$ & $\Conditioning$ & $\Target$ & $\Process$ & $\Features$ \\
\midrule
\multicolumn{6}{@{}l}{\textit{Classical settings}} \\
Classifier & $\log p(c \mid x)$ & --- & $c$ & $x \to f(x)$ & input features \\
\midrule
\multicolumn{6}{@{}l}{\textit{Autoregressive settings}} \\
Local next-token & $\log p(y_t \mid x, y_{<t})$ & --- & $y_t$ & $x \to y_1 \to \cdots \to y_t$ & prompt + prefix \\
Prompt-conditioned & $\log p(y_t \mid x, y_{<t})$ & $y_{<t}$ fixed & $y_t$ & $x \to y_1 \to \cdots \to y_t$ & prompt \\
Span-level prompt & $\log p(y_{1:T} \mid x)$ & $y_{1:T}$ fixed & $y_{1:T}$ & $x \to y_1 \to \cdots \to y_T$ & prompt \\
\midrule
\multicolumn{6}{@{}l}{\textit{Diffusion settings}} \\
State-level & $\log p(z_t \mid x, z_{>t})$ & --- & $z_t$ & $z_T \to \cdots \to z_0 \to y$ & prompt + generated states \\
Denoising-stage & $\log p(y \mid x) - \log p(y \mid x; \text{pert}(s_t))$ & --- & $y$ & $z_T \to \cdots \to z_0 \to y$ & denoising stages \\
Prompt-to-output & $\log p(y \mid x)$ & --- & $y$ & $z_T \to \cdots \to z_0 \to y$ & prompt \\
\bottomrule
\end{tabular}
\end{table}

\subsection{Autoregressive settings}
\label{subsec:Autoregressive_settings}

Autoregressive language models provide the most direct extension of classical feature attribution. Even in this familiar setting, feature attribution can be posed in different ways depending on whether the explanatory target is a next-token prediction, a prompt-conditioned token prediction, or a generated text span.

\paragraph{Local next-token attribution.}
Both prompt tokens and generated-prefix tokens are eligible, and nothing is held fixed. The explanatory question is local: which available context features helped the model predict the next token $y_t$?

\paragraph{Prompt-conditioned token attribution.}
This setting also explains a single target token, but restricts attribution to the prompt $x$ and holds the generated prefix fixed. The prefix still contributes to the prediction, though it is no longer eligible for attribution\footnote{The attributions sum to the prompt's contribution given the prefix, not to the model's total behavior at that step.}. The explanatory question is narrower: which prompt tokens explain the target token given the already-generated context?

\paragraph{Span-level prompt attribution.}
This setting shifts the target from a single token to a whole generated span $y_{1:T}$, with the prompt $x$ as the only eligible feature. The generated span is held fixed, and the score combines the conditional log-probability of each token in the span. The process $\Process$ is critical because $\log p(y_{1:T} \mid x) = \sum_t \log p(y_t \mid x, y_{<t})$ evaluates the prompt against the full autoregressive factorization rather than a single prediction step.

\paragraph{Self-attribution fallacy.}
Given the aforementioned settings, we identify an interpretation mistake in
feature attribution for language models. That is, reading a local next-token attribution
as though it came from the prompt-conditioned contract. 
We concur that attribution to generated-prefix tokens is not inherently wrong, and under a local next-token contract it is exactly what the method should recover. 
The mistake is to see most of the mass on the prefix and conclude that the prompt
did not matter.

A concrete scenario illustrates the failure mode. Consider a practitioner debugging an autoregressive language model that produced a hallucinated fact in a summary of a source document. The hallucinated token does not appear in the source. The practitioner runs a feature attribution method to explain the hallucinated token, with the model's input consisting of the source document followed by the partially-generated summary. The resulting attribution map shows that the immediately preceding generated tokens carry most of the mass, while the source document carries little. A natural reading is that \textit{the source document did not cause the hallucination, and that the model produced it from its own prior tokens}.

This reading commits the \emph{self-attribution fallacy}. The attribution was computed under a local next-token contract: $\Features$ is the full input including the generated prefix, $\Target$ is the hallucinated token, and $\Score$ is the log probability of that token. The map correctly identifies which tokens were predictive of the next-token prediction. It does not answer the question the practitioner would actually be asking: \emph{which features of the source document explain the hallucinated content of the summary?}. Answering that question requires a different contract: $\Features$ is the source document, $\Conditioning$ holds fixed the generated prefix up to the hallucinated text, $\Target$ is the hallucination itself.

The two contracts point at different fixes for the hallucination: one at the model's own decoding behavior, the other at a span of the source document. Naming the contract determines which fix the attribution can support.

Recent work on attribution for autoregressive language models has identified related instances of contract confusion: CAGE~\citep{walker2025cage} argues that current context-attribution methods give incomplete explanations because they ignore inter-generational influence, and HETA~\citep{pramanik2026heta} argues that linear approximations from classifier-era attribution fail to capture autoregressive structure. These are specific instances of the more general problem the Attribution Contract is designed to make explicit.

\subsection{Diffusion settings}

Diffusion language models include three main families: masked diffusion, uniform diffusion, and continuous diffusion models \citep{nie2025llada,lou2024sedd,jo2025continuous}. We use the term \emph{state} as a common abstraction across these families. In masked or uniform diffusion models, a state may correspond to a partially resolved token sequence. In continuous diffusion models, it may correspond to a continuous latent representation.
There is no fixed left-to-right ordering and no canonical next token to predict in diffusion LMs. Generation proceeds through an iterative process, and attribution can target a state of refinement, a denoising stage, or a final output.

An example illustrates what differs across these contracts. Consider a masked diffusion language model prompted with \texttt{The capital of France is} and generating an answer through a sequence of partially masked states. As generation proceeds, different positions are resolved until the final output contains \texttt{Paris}. This process gives rise to several feature-attribution questions:
\begin{itemize}
    \item \emph{State-level attribution} asks: what explains the model's prediction at a particular intermediate state, given the prompt and the states that came before it?
    \item \emph{Denoising-stage attribution} asks: which denoising stage was most responsible for committing the output to \texttt{Paris} rather than another capital?
    \item \emph{Prompt-to-output attribution} asks: which prompt features (e.g., \texttt{France}, \texttt{capital}) explain the final output \texttt{Paris}?
\end{itemize}

We now make each of these contracts precise.

\paragraph{State-level attribution.}
This setting targets an intermediate state $z_t$ rather than a final token, scored by $\log p(z_t \given x, z_{>t})$, where $z_{>t} = (z_{t+1}, \ldots, z_T)$ are the states committed earlier in the denoising chain. (Higher $t$ is earlier generation, the reverse of the autoregressive convention.) Both the prompt and the earlier states are eligible, and the question is which of them shaped the current state, so attribution here explains how an intermediate state evolves.

\paragraph{Denoising-stage attribution.}
Here the eligible features are the denoising stages, and the target is the final output $y$, scored by $\log p(y \mid x)$. Unlike autoregressive generation, the output score does not decompose naturally into a sum of per-stage scores. Attribution must therefore act on the stages directly. For example, the importance of stage $s_t$ can be measured by perturbing that stage and measuring the resulting change in $\log p(y \mid x)$. This setting asks which stages of the denoising process contributed most to the final output.

\paragraph{Prompt-to-output attribution.}
This setting targets the final output $y$, scored by $\log p(y \given x)$, with attribution restricted to the prompt $x$. It is the diffusion analog of span-level prompt attribution: the question is the same, which prompt features shaped the output, but the influence propagates through a denoising process $\Process$ instead of a left-to-right order.

These settings are not interchangeable. They answer different feature-attribution questions, so a single generic evaluation standard does not fit them all.

\subsection{Evaluating each contract}
\label{sec:eval}

One rule covers every contract we mentioned. A faithfulness test perturbs only the eligible features $\Features$ while the conditioning $\Conditioning$ stays fixed, and it reads the change in the contract's own score $\Score$, the column already listed in Table~\ref{tab:contracts}. For a token-level contract this would be deletion and insertion: remove or reveal the top-attributed features in attribution order and re-score under $\Score$~\citep{samek2016evaluating, madsen2023faithfulness}.

There are a few caveats. When the target is a span or a full output, the test perturbs the prompt and re-scores the \emph{original} generation. Regenerating a fresh sequence answers a different contract, and on a diffusion model that re-scoring runs the whole denoising chain. For denoising-stage attribution, the eligible features are stages, so a token perturbation has nothing to act on. There, one ablates a stage or perturbs its noise schedule and runs the chain to the end.

\section{Contract choices change the attribution}
\label{sec:empirical}

The contract does more than describe an attribution setting: its elements affect the resulting attribution. We test this directly by changing one element at a time while holding the model, method, and prompts fixed.

The Process element $\Process$ is different because it cannot be flipped within a single model. We therefore use two models: a mixture-of-experts model~\citep{muennighoff2024olmoe} for the within-model flips and a masked-diffusion model~\citep{team2026scaling} for the cross-process comparison. In our setups, discrete selection lies on the attribution path for the mixture-of-experts model and off the path for the masked-diffusion model.

\paragraph{Setup.} The mixture-of-experts (MoE) model is OLMoE-1B-7B~\citep{muennighoff2024olmoe}; the masked-diffusion model is Steerling-8B~\citep{team2026scaling}, used for the Process experiment in Section~\ref{sec:cond-flip}. Attribution is embedding-injection Integrated Gradients on the sampled-token logit, integrated from a zero (all-zeros embedding) baseline with right-Riemann quadrature~\citep{sundararajan2017axiomatic}, over greedy-generated answers of up to 16 tokens. The mixture-of-experts model runs in bfloat16 on a single A100 GPU; the prompts are the first 120 questions from the validation split of Natural Questions Open~\citep{kwiatkowski2019natural}, a public question-answering dataset. All reported means and
correlations are over these 120 questions unless noted otherwise.
Where a flip calls for a disagreement measure, we report the Spearman rank correlation and the top-5 overlap of the two prompt-token rankings; a low value means the element reordered the ranking. 
Unless noted otherwise, experiments are done on the MoE model.

\paragraph{Integrated Gradients.} Integrated Gradients (IG)~\citep{sundararajan2017axiomatic} attributes a scalar output $f(x)$ to the inputs by integrating the gradient along the
straight-line path from a baseline $x_0$ to the input $x$: $A_i = (x_i - x_{0,i}) \int_0^1 \partial f(x_0 + \alpha(x - x_0))/\partial
x_i \, d\alpha$, approximated by a sum over $n$ quadrature steps. 
Integrated Gradients satisfies the completeness identity
\[
\sum_i A_i = f(x) - f(x_0)
\]
for sufficiently smooth functions along the integration path.
For a language model, the input is the sequence of token embeddings: we interpolate from a zero baseline embedding
to the real embeddings, and $f$ is the logit of the sampled token. The interpolation parameter $\alpha$ traces the path to which we refer 
throughout this section, and the gradient at each $\alpha$ is the integrand.

\paragraph{Completeness.} Completeness measures how much of the output change the attribution accounts for. The residual
\[
\left\lvert \sum_i A_i - (f(x) - f(x_0)) \right\rvert
\]
is the part of the score change the attribution does not capture. We aggregate over generated target tokens by mass weighting:
\[
\frac{
\sum_t \left\lvert \sum_i A_i^{(t)} - \bigl(f_t(x)-f_t(x_0)\bigr) \right\rvert
}{
\sum_t \left\lvert f_t(x)-f_t(x_0) \right\rvert
}.
\]
This gives more weight to targets with larger score changes and prevents targets with near-zero changes from inflating the ratio. Zero means exact completeness.
The ratio can exceed $100\%$. When a coarse quadrature overshoots the score change, a target's numerator can exceed its denominator, so residuals above 100\% appear at very low step counts.


\subsection{Features: the eligible set $\Features$ changes the attribution}
\label{sec:feat-flip}

Under the local next-token contract, the generated prefix is an eligible feature, and it takes a large share of the attribution.
Across the Natural Questions (NQ) prompts, the generated prefix carries 29\% of the absolute
attribution mass on average, close to the question at 25\% and well above the system prompt at 10\%.
The chat-template tokens take the remaining 36\%, much of it on the BOS token, which acts as an attention sink~\citep{xiao2024efficient}
and absorbs mass without carrying content.
A reader who interprets this map as answering ``which prompt tokens explain the output'' would misread it, since almost as much mass sits on the model's own generated tokens as on the question. This is the self-attribution fallacy from Section~\ref{sec:settings}, and here we can measure its extent.

Making the prefix ineligible changes the conclusion, not only the totals. We rerun the same attribution under the prompt-conditioned contract, which holds the generated prefix fixed and attributes only the prompt. We implement this with a per-position baseline whose generated positions equal their real embeddings, so the prefix stays off the integration path and receives exactly zero attribution. The prompt-token ranking then shifts: its Spearman rank correlation with the local ranking is $0.63$, with a top-5 overlap of $0.56$ (Table~\ref{tab:flips}), meaning that about half of the top prompt tokens change. Which prompt tokens receive the most attribution therefore depends on whether the prefix is eligible.

\subsection{Conditioning $\Conditioning$ decides whether attribution completeness is satisfied}
\label{sec:cond-flip}

\paragraph{Mixture-of-experts routing.} A mixture-of-experts layer replaces one feedforward block with many expert blocks and routes each token to
a few of them. A router scores the experts for the token and keeps the top $k$; the token is processed by those $k$ experts and
combined by continuous gate weights read from the same scores. Two things therefore vary with the input: the discrete selection of
which experts, and the continuous weights of how much each contributes.

Our experiment changes this discrete selection, which is a conditioning choice in our contract. At each interpolation step the router either recomputes the selection, or replays the selection it made at the real input, in which case only the continuous gate weights change along the path. Both attribute the same target on the same model, and differ only in whether the discrete selection is recomputed or held fixed.

\begin{figure}[!hb]
\centering
\includegraphics[width=0.5\linewidth]{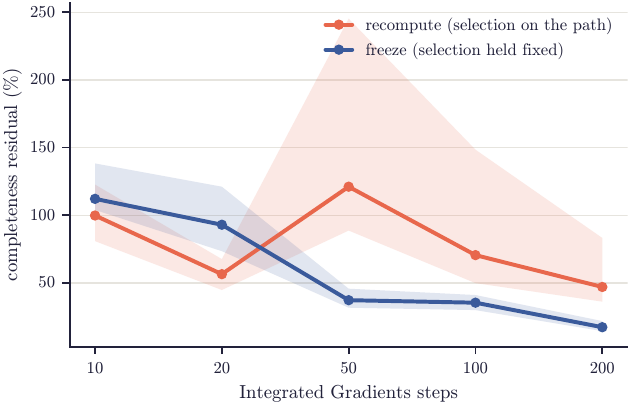}
\caption{\textbf{Freezing the router selection lowers the completeness residual.} Completeness residual against Integrated Gradients steps on the mixture-of-experts model, over 60 questions, median with an inter-quartile band. When the selection is recomputed, the residual stays high and unstable; when it is held fixed, the residual descends and sits below the recomputed one from $n{=}50$ on.}
\label{fig:cond}
\end{figure}

Figure~\ref{fig:cond} gives the completeness residual across steps under the two settings, over 60 questions. The two settings separate, which identifies the router selection as the dominant source of the residual. Recomputed, the residual is large and unstable: its median moves non-monotonically, $56\%$ at $n{=}20$, $121\%$ at $n{=}50$, $47\%$ at $n{=}200$. Freezing the router selection changes this. From $n{=}50$ on, the frozen residual falls below the recomputed one on nearly every question (60/60 at $n{=}50$, 57/60 at $n{=}200$), and its median descends smoothly from $37\%$ at $n{=}50$ to $17\%$ at $n{=}200$. Discrete expert selection sits on the integration path and re-runs inside the attributed forward pass, injecting output changes that carry no gradient. Freezing it removes most of that effect. The frozen residual does not reach zero, so freezing does not restore exact completeness, but it removes most of the gap.

We piloted the same experiment on two other public mixture-of-experts models, Qwen1.5-MoE-A2.7B~\citep{qwen_moe} and Mixtral-8x7B~\citep{jiang2024mixtral}, and saw the same qualitative behavior: the recomputed residual is large and non-monotonic, and freezing the selection moves it substantially without reaching a clean floor. 
How clean this contrast looks is architecture-dependent. On other mixture-of-experts architectures the two settings separate sharply, with a stable recomputed floor and a near-zero frozen residual, while on the softmax-routed models here neither setting settles to a low floor. What reproduces across architectures is: the discrete routing selection is where the completeness failure comes from.

To rule out an implementation bug, we check a case where IG is exact. On a linear model, our implementation satisfies completeness to a relative deviation below $10^{-5}$ and recovers the exact attributions $W(x - x_0)$. So the residual on the mixture-of-experts model is a property of the model on the integration path, not a defect of IG.

The Conditioning element $\Conditioning$ therefore strongly affects feature attribution in this setting. Recomputing the router selection leaves a large, unstable residual; holding it fixed removes most of it. The Process $\Process$ element acts through this choice: a mixture-of-experts model recomputes its routing on the path, while a masked-diffusion model fixes its unmasking order at generation and replays it, so the process sets the default conditioning.

A second model supports the process $\Process$'s importance. On Steerling~\cite{team2026scaling}, the selection is fixed by construction, and completeness behaves in the opposite way: the residual converges with steps. 
On the Base model it falls from $26\%$ at $n=128$ to $18\%$ at $n=256$. On the Instruct variant it reaches $13\%$ at $n=2048$, with small increases at $n=32$ and $n=512$ along the way. The curve does not flatten over the range we run (Table~\ref{tab:diffusion}). 

\begin{table}[!h]
\centering
\small
\setlength{\tabcolsep}{6pt}
\begin{threeparttable}
\caption{\lead{Diffusion completeness residual generally decreases with more integration steps.} Normalized residual ($\%$) vs. IG steps.}
\label{tab:diffusion}
\begin{tabular}{@{}lcccccccc@{}}
\toprule
\headrow steps & 1 & 8 & 32 & 128 & 256 & 512 & 1024 & 2048 \\
\midrule
Base & 97 & 88 & 47 & 26 & 18 & --- & --- & --- \\
Instruct & 128 & 72 & 74 & 27 & 21 & 24 & 15 & 13 \\
\bottomrule
\end{tabular}
\end{threeparttable}
\end{table}


We do not run the diffusion model to full convergence, so we do not compare the two models by the size of their residuals. At the step
counts we can afford, the diffusion residual sits in the same broad range as the mixture-of-experts residual. The difference that reproduces is
the shape of the curve as steps increase: the diffusion residual generally decreases, while the mixture-of-experts residual stays high and unstable.

This comparison is across two models rather than a within-model flip like the frozen-routing result, so it supports the role of the
generative process without isolating it. 
The finding also bears on \citet{bilodeau2024impossibility}, whose theorem applies to methods that are complete and linear, such as Integrated Gradients. On the mixture-of-experts model, Integrated Gradients is not complete, so it falls outside that hypothesis class. It narrows where the impossibility theorem applies, and shows that whether a method meets the precondition depends on the generative process.

\subsection{Score $\Score$: the attributed quantity changes the attribution ranking}
\label{sec:score-flip}

Attributing a different scalar $\Score$ of the same token $\Target$ changes the ranking. We attribute the sampled token's logit, its log-probability, and its probability, with everything else fixed, and compare the prompt-token rankings. 
The three scores refer to the same target token but define different scalar functions of the model output. Although probability and log-probability preserve the ranking of output tokens, attribution depends on input gradients, which differ across these scores and from the raw logit.
The raw logit depends only on the sampled token; the log-probability normalizes over
the vocabulary, so its gradient subtracts a probability-weighted average of all token logits; the probability rescales that gradient by the token's own probability. The input-token rankings therefore differ even though the output ranking does not. The logit and the log-probability agree at Spearman $0.57$, the logit and the probability at $0.62$, and the log-probability and the probability at
$0.47$, with top-5 overlaps between $0.43$ and $0.60$. No pair matches, so the Score element $\Score$ is influential (Table~\ref{tab:flips}).

\begin{table}[!h]
\centering
\small
\setlength{\tabcolsep}{8pt}
\begin{threeparttable}
\caption{\lead{No score or conditioning change preserves the ranking.} Spearman correlation and top-5 overlap of the prompt-token rankings lower means a larger effect.}
\label{tab:flips}
\begin{tabular}{@{}llcc@{}}
\toprule
\headrow Flip & Comparison & Spearman & top-5 overlap \\
\midrule
Score & logit vs log-prob & 0.57 & 0.60 \\
Score & logit vs prob & 0.62 & 0.54 \\
Score & log-prob vs prob & 0.47 & 0.43 \\
Conditioning & local vs prompt-conditioned & 0.63 & 0.56 \\
\bottomrule
\end{tabular}
\end{threeparttable}
\end{table}

\subsection{Evaluation must match the contract}
\label{sec:eval-flip}

That completeness fails on the MoE model does not make its attributions useless. Under the deletion and insertion tests that fit the local next-token contract, the attribution ranking outperforms random rankings on 91\% of targets under deletion and 78\% under insertion (the area-under-curve gap against a random ordering, over $k{=}12$ perturbed positions and five random orders). The completeness residual and the faithfulness gap are moreover at most weakly correlated: their Spearman correlation is $-0.26$ for deletion and $-0.22$ for insertion. The axiom residual carries little information about whether the ranking is useful~\citep{bilodeau2024impossibility}. Judging this contract by its completeness axiom would reject rankings that its faithfulness test shows to be correct, the mismatch Section~\ref{sec:eval} warns against.



\section{Related work}

\subsection{Classical feature attribution}

Classical feature attribution methods assign importance scores to input features relative to a model output, with Integrated Gradients~\citep{sundararajan2017axiomatic} as a canonical method motivated by axioms. Subsequent work has shown that even in classical settings, attribution can be sensitive to the downstream task and evaluation criterion \citep{nguyen2021effectiveness, adebayo2022post}. 
One important choice is what to hold fixed: \citet{chen2020true} show that computing a feature's attribution by intervening on the remaining features rather than conditioning on them gives different attributions for the same model and input.
In a generative language model, what is held fixed can also be the model's own output, or a discrete internal decision such as expert routing.
\citet{bilodeau2024impossibility} go further: they prove that complete and linear attribution methods, including Integrated Gradients~\citep{sundararajan2017axiomatic} and SHAP~\citep{lundberg2017unified}, can fail to outperform random guessing on concretely specified end-tasks such as spurious-feature identification and algorithmic recourse. Their result presumes the task is already fixed. 
We ask: in a generative language model, what does fixing the task really consist of?


The broader argument that interpretability claims are under-specified without explicit attention to their motivations and properties goes back to \citet{lipton2017mythos}. For attribution, this under-specification appears on two axes. The first is computational: Integrated Gradients interpolates in a straight line between embeddings, a path that passes through points corresponding to no real token, and Discretized Integrated Gradients~\citep{sanyal2021dig} and Sequential Integrated Gradients~\citep{enguehard2023sig} show that a different path changes the scores. The second axis is the question itself. \citet{jacovi2020towards} argue that faithfulness has no single definition and that an evaluation is meaningful only relative to the one it assumes, and \citet{meloux2025deadsalmons} frame the general version as non-identifiability: many explanations are consistent with the same computational traces unless the query, hypothesis class, and discrepancy measure are specified. The Attribution Contract works on this second axis. Generative language models widen it, since the explanatory target, the eligible feature set, the conditioning regime, and the generation process can all vary independently, and Section~\ref{sec:eval} gives the consequence for evaluation: a faithfulness test must perturb the eligible features and re-score the quantity the contract names.



\subsection{Feature attribution for generative language models}

Several recent works extend feature attribution to sequence generation and decoder-only language models. Inseq~\citep{sarti2023inseq} provides an interpretability toolkit for sequence generation, Captum's generative-LM interface~\citep{miglani2023captum} adapts attribution tooling to generated text, and ReAGent~\citep{zhao2024reagent} proposes a model-agnostic method for generative LMs. More recent work such as HETA~\citep{pramanik2026heta} and CAGE~\citep{walker2025cage} targets autoregressive decoder-only generation, including higher-order effects and inter-generational influence among generated tokens. ContextCite~\citep{cohenwang2024contextcite} attributes generation to context features, and Jacobian Scopes~\citep{liu2026jacobian} propose three gradient-based variants targeting different model predictive outputs.

These methods are often read as competing alternatives. We argue they are not.
ContextCite operates within a prompt-to-output contract: its target is the final generation, its eligible features are context sources, and its score is the joint sequence probability. CAGE does not, since its eligible set also includes all prior generations, placing it closer to a span-level contract with the prefix eligible. Inseq and Captum typically operate within local next-token contracts: their target is a single next-token prediction, their eligible features include the generated prefix, and their score is the next-token log-probability. HETA argues that linear attribution methods miss higher-order effects across generation steps, which is a within-contract critique of a particular method family rather than a competing alternative. Jacobian Scopes apply gradient-based attribution to three different scores: a specific logit (Semantic), the full predictive distribution (Fisher), and model confidence (Temperature). These variants use the same local next-token contract and the same underlying attribution algorithm; they differ only in the score element $\Score$. The Attribution Contract makes this classification explicit: methods that look comparable on the surface may answer different questions, and methods that look different may answer the same one.

The score element has been varied before: \citet{yin2022interpreting} show that explaining why a model produced token $a$ rather than a contrastive alternative $b$ gives different and more interpretable attributions than explaining $a$ alone. Section~\ref{sec:score-flip} makes a finer point, that even monotone transformations of a single target score, which represent the output identically, reorder the input tokens.

\section{Limitations}

\paragraph{The contract taxonomy is not exhaustive.}
SCOPE is intended to specify an attribution question, not to enumerate every possible attribution setting. The contracts in Section~\ref{sec:settings} illustrate common autoregressive and diffusion settings, but other targets, conditioning choices, processes, and feature sets are possible.

\paragraph{Contract choice remains application-dependent.}
The Attribution Contract specifies the explanatory question an attribution answers, but it does not prescribe which contract should be used for every application. That choice depends on the explanatory goal.
 
\paragraph{Scope of the measurements.}
Our experiments use 120 prompts from Natural Questions Open with greedy generations of up to 16 tokens, so the mass split in Section~\ref{sec:feat-flip} is a measurement on short-form question answering. The 120 prompts are a small set of the $3{,}610$-question NQ validation set, chosen for compute budget; the routing-floor experiment, which sweeps many integration steps, uses 60 of them. The sample size reflects our compute budget rather than a limitation of the experimental pipeline.

\paragraph{Method and models.}
Every result uses Integrated Gradients. The attribution completeness study is specific to it by construction. We mainly use two models, an autoregressive mixture-of-experts model and a masked-diffusion model, chosen so the discrete selection lies on opposite sides of the integration path. Extending the study to other architectures and methods is future work.
 
\paragraph{The Process comparison is not isolated.}
The conditioning result is a within-model flip. The process comparison is not, since the two
models differ in architecture, scale, and training data alongside the property of interest. We
therefore compare the shape of the residual curve and not its magnitude.

\section{Conclusion}

Feature attribution in generative language models cannot be treated as a straightforward extension of classifier-era attribution. In classical settings, it is often acceptable to assume a stable input-output structure and a fixed feature set. In generative language models, these assumptions break down.

This paper introduced the \emph{Attribution Contract}, the explanatory setting under which a feature-attribution claim is made.
We used this framework to distinguish several feature-attribution settings in generative language models. We also identified a common interpretive error in autoregressive models, the \emph{self-attribution fallacy}: reading attribution to generated-prefix tokens as if it answered a prompt-level question.

Empirically, we showed that the Conditioning element shapes whether an attribution method can satisfy its own axioms. On a mixture-of-experts model, recomputing discrete routing along the integration path leaves Integrated Gradients with a large, unstable completeness residual, while holding the routing fixed lowers it substantially and identifies the routing selection as the source. The Process element appears to set which conditioning a model imposes by default: a mixture-of-experts model recomputes routing on the path, whereas a masked-diffusion model fixes its unmasking order at generation and replays it, and the diffusion residual generally decreases where the mixture-of-experts residual does not. That comparison is across two models rather than a within-model flip, so it indicates the role of the process without isolating it.

Completeness failure does not make an attribution useless. On the mixture-of-experts model the same rankings pass deletion and insertion under the contract they were computed for, and the completeness residual carries little information about whether they do. The broader implication is that feature-attribution methods in generative language models cannot be evaluated against a single standard. Different contracts ask different questions, evaluation must match the contract, and methods should therefore be evaluated as method-contract pairs.

\section*{Reproducibility Statement}

All experiments use publicly available models and data: OLMoE-1B-7B, Steerling-8B, and
the validation split of Natural Questions Open. Section~\ref{sec:empirical} gives the attribution
configuration in full: embedding-injection Integrated Gradients on the sampled-token logit, a
zero baseline, right-Riemann quadrature, greedy decoding to at most 16 tokens, in bfloat16 on a
single A100. The two conditioning regimes differ only in whether the router recomputes its top-$k$
selection at each interpolation step or replays the selection made at the real input. We implement
the frozen condition by caching the router's selection indices during the real-input forward pass and
substituting them at every subsequent step, leaving the continuous gate weights free to vary. As a
correctness check, our implementation recovers the exact Integrated Gradients attributions
$W(x - x_0)$ on a linear model to a relative deviation below $10^{-5}$. 

\section*{Acknowledgments}

The author thanks Aya Abdelsalam Ismail, Anh Totti Nguyen, and Julius Adebayo for their comments on earlier drafts of this paper. The views expressed in this paper are those of the author and do not necessarily reflect the views of Guide Labs.

\bibliography{references}
\bibliographystyle{tmlr}

\end{document}